\documentclass{article}

\usepackage{microtype}
\usepackage{graphicx}
\usepackage{subfigure}
\usepackage{booktabs} %
\usepackage{natbib}
\usepackage{multirow}
\usepackage{rotating}
\usepackage{enumitem}
\usepackage{tablefootnote}
\usepackage{manyfoot}
\DeclareNewFootnote{affil}

\usepackage{hyperref}

\usepackage[accepted]{icml2025}

\setcitestyle{numbers,sort&compress,open={[},close={]}}
\hypersetup{
colorlinks=false,
linkbordercolor=cyan,
citebordercolor=green,
urlbordercolor=cyan,
pdfborder={0 0 1}
}

\usepackage{amsmath}
\usepackage{amssymb}
\usepackage{mathtools}
\usepackage{amsthm}
\usepackage{amsfonts}       %
\usepackage{xspace}
\usepackage{xpatch}
\usepackage{xcolor}
\usepackage{wrapfig}
\usepackage{booktabs}       %

\usepackage{array}
\newcolumntype{L}[1]{>{\raggedright\arraybackslash}m{#1}}

\theoremstyle{plain}

\theoremstyle{definition}

\theoremstyle{remark}

\usepackage[textsize=tiny]{todonotes}

\newcommand{\name}{\textsc{Adver\-saria\-LLM}\xspace}
\newcommand{\judgezoo}{\textsc{JudgeZoo}\xspace}

\icmltitlerunning{Preprint.}

\begin{document}

\twocolumn[
\icmltitle{AdversariaLLM: A Unified and Modular Toolbox for LLM Robustness Research}

\icmlsetsymbol{equal}{*}
\begin{icmlauthorlist}
\icmlauthor{Tim Beyer}{cs,mdsi}
\icmlauthor{Jonas Dornbusch}{cs,mdsi}
\icmlauthor{Jakob Steimle}{cs,mdsi}
\icmlauthor{Moritz Ladenburger}{cs,mdsi}
\icmlauthor{Leo Schwinn}{cs,mdsi}
\icmlauthor{Stephan G\"unnemann}{cs,mdsi}
\end{icmlauthorlist}

\icmlaffiliation{cs}{Department of Computer Science, Technical University of Munich, Germany}
\icmlaffiliation{mdsi}{Munich Data Science Institute, Germany}

\icmlcorrespondingauthor{Tim Beyer}{tim.beyer@tum.de}
\icmlkeywords{Machine Learning, ICML}
\vskip 0.3in
]

\printAffiliationsAndNotice

\begin{abstract}
  The rapid expansion of research on Large Language Model (LLM) safety and robustness has produced a fragmented and oftentimes buggy ecosystem of implementations, datasets, and evaluation methods. This fragmentation makes reproducibility and comparability across studies challenging, hindering meaningful progress. To address these issues, we introduce \href{https://github.com/LLM-QC/AdversariaLLM}{\name}\footnote{\url{https://github.com/LLM-QC/AdversariaLLM}}, a toolbox for conducting LLM jailbreak robustness research. Its design centers on reproducibility, correctness, and extensibility. The framework implements twelve adversarial attack algorithms, integrates seven benchmark datasets spanning harmfulness, over-refusal, and utility evaluation, and provides access to a wide range of open-weight LLMs via Hugging Face. The implementation includes advanced features for comparability and reproducibility such as compute-resource tracking, deterministic results, and distributional evaluation techniques. \name also integrates judging through the companion package \href{https://github.com/LLM-QC/judgezoo}{\judgezoo}\footnote{\url{https://github.com/LLM-QC/judgezoo}}, which can also be used independently. Together, these components aim to establish a robust foundation for transparent, comparable, and reproducible research in LLM safety.
\end{abstract}

\section{Introduction}
Despite rapid growth in LLM robustness research, including new attacks, defenses, and benchmarks, evaluation practices remain underdeveloped.
A key reason is the highly fragmented LLM evaluation ecosystem: numerous jailbreak attack, defense, and benchmark implementations diverge in subtle but consequential ways \citep{UK_AI_Security_Institute_Inspect_AI_Framework_2024,chao2024jailbreakbench,garak,mazeika2024harmbench,meta_llama_purplellama_2025,munoz2024pyritframeworksecurityrisk,shen2025pandaguard,wolf2024tradeoffs,zhu2023promptbench}.
Implementations often contain bugs, inconsistently defined threat models, non-standard dataset splits, and undisclosed or widely varying judging criteria.
These factors undermine comparability and reproducibility across studies. \citep{beyer2025llm,rando2025adversarial}

In line with recent demands for more consistency in LLM safety evaluations \citep{beyer2025llm,rando2025adversarial,schwinn2025adversarial}, we introduce \name, a unified, modular toolbox for LLM robustness research designed around three principles: reproducibility, correctness, and extensibility.
Alongside \name, we release \judgezoo, a companion package that standardizes evaluation by implementing 13 judges from prior work in a tested, reproducible setup.

Our key contributions are:
\begin{itemize}[leftmargin=*,noitemsep]
    \item \textbf{Corrected implementations}: We identify and fix critical bugs in tokenization filtering, chat templates, and batched generation that affect existing tools, demonstrating up to 28\% ASR improvement from correctness fixes alone (Figure~\ref{fig:whitespace-token}).
    \item \textbf{Comprehensive coverage}: 12 attack algorithms spanning discrete, continuous, and hybrid methods; 6 datasets covering harmfulness and over-refusal; 13 evaluation judges.
    \item \textbf{Advanced features}: Budget tracking (queries, time, FLOPs), per-step evaluation, distributional assessment via Monte Carlo sampling, and 2.12× more consistent batched generations.
    \item \textbf{Standardized evaluation}: \judgezoo makes it easy to reproduce evaluation setups from prior work and warns when configurations deviate from established baselines.
\end{itemize}

We release both packages publicly and hope they can help establish more rigorous, comparable standards for LLM safety research.

\begin{figure*}[t]
  \centering
  \includegraphics[width=0.8\linewidth]{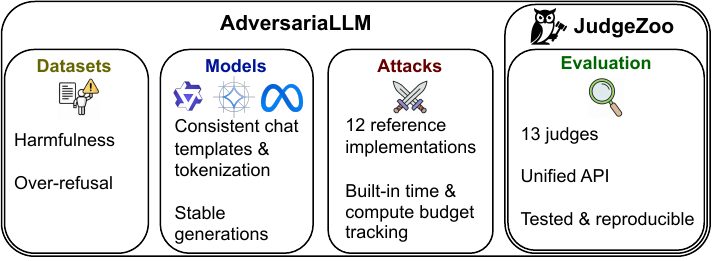}
  \caption{\name is a framework for reproducible and principled LLM adversarial robustness evaluation.}
  \label{fig:hero}
\end{figure*}

\section{Related work}

\paragraph{Safety \& Over-refusal Datasets}
Recent benchmark suites like HarmBench~\citep{mazeika2024harmbench}, JailbreakBench~\citep{chao2024jailbreakbench}, and StrongREJECT~\citep{souly2024strongreject} aim to systematize harmfulness evaluation by defining threat models, reference datasets, and consistent scoring protocols.
Complementary work like XSTest~\citep{rottger2024xstest} and OR-Bench~\citep{cui2024or} address the converse failure mode by measuring over-refusal on benign prompts.
Together, these datasets provide a foundation for evaluating safety–utility trade-offs.
However, their implementations have remained scattered and difficult to adapt, which creates subtle but critical evaluation challenges.
For instance, an adversarially trained model using system message A during training may appear robust yet over-cautious in that configuration, but when evaluated with a different system message B, safety characteristics and over-refusal might be reduced, undermining fair comparison.
Our modular design enables evaluations with \name to remain consistent across datasets and models.

\paragraph{Tooling for adversarial evaluation of LLMs}
In computer vision, libraries like CleverHans \citep{papernot2018cleverhans} and RobustBench \citep{croce2020robustbench} advanced the field by establishing consistent standards for adversarial robustness evaluation and providing ways to reliably measure progress.
In the LLM space, initial attacks were often built entirely in bespoke repositories \citep{zou2023universal}. However, over the last few years, different frameworks aiming to collect datasets, attack algorithms, and/or evaluation have emerged \citep{UK_AI_Security_Institute_Inspect_AI_Framework_2024,chao2024jailbreakbench,garak,mazeika2024harmbench,meta_llama_purplellama_2025,munoz2024pyritframeworksecurityrisk,shen2025pandaguard,wolf2024tradeoffs,zhu2023promptbench}.
Nevertheless, subtle implementation differences and bugs, such as incorrect chat templates or insufficient filtering of unreachable token sequences, can lead to divergent results.
In addition, many existing frameworks (e.g., HarmBench \citep{mazeika2024harmbench}) focus exclusively on adversarial attacks, overlooking over-refusal evaluation, despite it being crucial for fairly comparing defenses and safety training approaches \citep{an2024automatic,beyer2025llm,thompson2024breaking}.
\name seeks to resolve these issues by providing comprehensive and correct implementations of existing baselines, enabling faithful comparison across datasets, models, attacks, and judges. Its modular design encourages adoption by making experimentation and expansion simple.
\begin{table*}[h]
\centering
\label{tab:framework_comparison}
\begin{tabular}{lccccc}
\toprule
Framework & 
\begin{tabular}[c]{@{}c@{}}Harmfulness\\Evaluation\end{tabular} & 
\begin{tabular}[c]{@{}c@{}}Over-Refusal\\Evaluation\end{tabular} & 
\begin{tabular}[c]{@{}c@{}}Embedding-space\\Attacks\end{tabular} & 
\begin{tabular}[c]{@{}c@{}}Attack Budget\\Tracking\end{tabular} & 
\begin{tabular}[c]{@{}c@{}}Distributional\\Evaluation\end{tabular} \\
\midrule
StrongREJECT~\citep{souly2024strongreject} & \checkmark & \texttimes & \texttimes & \texttimes & \texttimes \\

garak~\citep{garak} & \checkmark & \texttimes & \texttimes & (\checkmark)\tablefootnote{Only wall-time tracking and only for an entire run} & \texttimes \\
HarmBench~\citep{mazeika2024harmbench} & \checkmark & \texttimes & \texttimes & \texttimes & \texttimes \\
PandaGuard~\citep{mazeika2024harmbench} & \checkmark & \texttimes & \texttimes & \texttimes & \texttimes \\
\midrule
\name & \checkmark & \checkmark & \checkmark & \checkmark & \checkmark \\
\bottomrule
\end{tabular}
\caption{Comparing existing evaluation frameworks for LLM adversarial robustness evaluation with \name.}
\end{table*}

\paragraph{Evaluation and Judging}
Automated LLM-as-a-judge evaluation has become the de facto standard due to its low cost and scalability.
Currently, a wide variety of proposed techniques are in use~\citep{andriushchenko2024jailbreaking,chao2025jailbreaking,ghosh2024aegis,hughes2024best,li2024salad,liu2024jailjudge,llama2024llama3herd,mazeika2024harmbench,llama_guard_4,souly2024strongreject,zhu2024advprefix}.
Despite many available options, fine-tuned judge LLMs either offer poor alignment with human judgment, or their implementations contain bugs or lack scalability, while results from prompted foundation models remain hard to compare across works due to differences in prompt phrasing or model choice.
These inconsistent scoring schemes limit comparability, and motivate our standardized, modular judging approach, which makes evaluation setups reproducible and transparent.

\section{A unified, modular package for LLM robustness research}

We now describe the design of our framework in more detail.
\name is a framework designed for evaluating LLM robustness under adversarial conditions with the goal of making models, defenses and attacks easily and fairly comparable.
The framework is organized into four modules: \hyperref[subsec:datasets]{datasets}, \hyperref[subsec:models]{models}, \hyperref[subsec:attacks]{attack algorithms}, and \hyperref[subsec:evaluation]{evaluation}.
All modules communicate via standardized APIs and are easy to extend.
In the following sections we describe each component in more detail.

\subsection{Datasets}
\label{subsec:datasets}
\name provides a unified interface for six datasets across multiple dimensions of model behavior.
All datasets have consistent APIs using the de facto standard role and turn-based \texttt{[{"role": "<role>", "content": "<content>"}, {...}, ...]} conversations format, support deterministic shuffling and splits via controllable random seeds, and offer category-level filtering where applicable, enabling fine-grained and reproducible analysis across harm types and use cases.

\textbf{Adversarial robustness.} To test model safety when confronted with harmful instructions, we include HarmBench Behaviors \citep{mazeika2024harmbench}, which contains 520 harmful prompts spanning six risk categories; JailbreakBench \citep{chao2024jailbreakbench}, featuring 100 harmful prompts; and both small and full versions of the StrongREJECT dataset (60 and 313 prompts, respectively) \citep{souly2024strongreject}, comprising manually curated harmful prompts across six distinct harm categories.

\textbf{Over-refusal detection.} Over-refusal evaluations aim to measure false-positive refusals on benign content which may superficially look harmful.
We incorporate XSTest \citep{rottger2024xstest}, a manually curated set of 250 safe prompts designed to expose overly cautious refusal behavior, and OR-Bench-Hard-1k \citep{cui2024or}, containing 1,000 automatically generated safe prompts which are likely to trigger false refusals on frontier models.

\textbf{Other.} Given increasing interest in mechanistic interpretability in adversarial settings \citep{arditi2024refusal,ball2024understanding,wollschlager2025geometry} and recent attacks based on targeted manipulations of model weights or activations via refusal direction ablation, we also implement an interface to a dataset which provides both harmful and harmless prompts and can be used to compute refusal directions \citep{arditi2024refusal}.

\subsection{Models}
\label{subsec:models}

\name leverages the widely-used Hugging Face \citep{wolf2019huggingface} package for model implementations.
While this enables a high degree of flexibility and interoperability, we find that some of the provided implementations contain inconsistencies.
Common problems include:
\begin{itemize}[topsep=0em, parsep=0em, itemsep=2pt]
    \item Non-standard chat templates that deviate from model developers' defaults.  %
    \item Incorrect or incomplete padding and end-of-turn/end-of-text tokens.
    \item Inconsistent trimming of long sequences due to incorrect model context length settings.
\end{itemize}
To fix these issues, we use corrected templates from \citet{zheng2024chattemplates} where available, and manually fix special tokens and long sequence behavior for all supported models.
We also force practitioners to make explicit decisions about quantization and model weight data formats, as these factors can affect model robustness \citep{beyer2025llm}.

\subsection{Attacks}
\label{subsec:attacks}

\name implements adversarial attacks in discrete, continuous, and hybrid input spaces across both single- and multi-turn scenarios (see \autoref{tab:attacks}).

The existing suite can be easily extended with custom attack implementations:
An attack is defined as accepting a \texttt{Config} object with the relevant hyperparameters, and exposing a \texttt{run} method which receives a tuple consisting of a model, a tokenizer, and a dataset instance, and returns an \texttt{AttackResult} object.
These output objects are saved using a human-readable, JSON-based format to allow quick manual inspection (see \autoref{app:attack-result}).

\begin{table}[h]
\centering
\setlength{\tabcolsep}{2pt}
\small
\begin{tabular}{llp{2.2cm}p{4.2cm}}
\toprule
& Type & Attack & Description \\
\midrule
\multirow{6}{*}{\rotatebox{90}{Multi-turn}} 
  & Discrete & ActorAttack \citep{ren2024llms} & Uses prompts inspired by actor-network theory \\
  & Discrete & Crescendo \citep{russinovich2025great} & Black-box multi-turn jailbreak leveraging the victim model's replies to escalate into prohibited content \\
\midrule
\multirow{23}{*}{\rotatebox{90}{Single-turn}} 
  & Discrete & AmpleGCG \citep{liao2024amplegcg} & Transfer-based GCG with specialized prompter model \\
  & Discrete & AutoDAN \citep{liu2023autodan} & Genetic algorithm with crossover and mutation \\
  & Discrete & BEAST \citep{sadasivan2024fast} & Efficient fluent adversarial suffix generation via beam search \\
  & Discrete & Best-of-N \citep{hughes2024best} & Low-level perturbations with high query volume \\
  & Discrete & Direct & Baseline without adversarial optimization \\
  & Discrete & GCG \citep{zou2023universal} & Greedy coordinate gradient with token-level optimization \\
  & Discrete & HumanJailbreaks \citep{shen2024anything} & Curated human-written jailbreaks \\
  & Discrete & PAIR \citep{chao2025jailbreaking} & Multi-turn refinement with attacker LLM \\
  & Discrete & REINFORCE-GCG \citep{geisler2025reinforce} & Policy gradient variant using judge feedback \\
  & Cont. & PGD \citep{schwinn2024soft} & Gradient descent in embedding space with flexible norm constraints and optimization strategies \\
  & Hybrid & PGD-Discrete \citep{geisler2024attacking} & Continuous relaxation with repeated projection \\
\bottomrule
\end{tabular}
\caption{Overview of the adversarial attacks in \name, grouped by scenario with rotated labels.}
\label{tab:attacks}
\end{table}

\subsection{Evaluation}
\label{subsec:evaluation}
Alongside \name, we introduce \judgezoo, a companion package, focused on consistent, automated evaluation of potentially harmful model generations.
While uptake of prior LLM adversarial attack frameworks has so far been limited, with new work (e.g., \citep{chao2025jailbreaking,sadasivan2024fast,schwinn2024soft,zou2023universal}) often relying on purpose-built repositories rather than integrating with existing toolkits, we see judging as having a particularly high chance to be unified and modularized.
For this reason, \judgezoo is designed to function independently of \name.

As shown in \autoref{tab:judges}, \judgezoo implements over 12 different judges and provides a unified API for scoring model outputs (e.g., probability of harm, discrete ratings or specific harm categories).
\judgezoo supports both fine-tuned models and prompt-based judging, and allows the user to switch between local models or remote foundation models from providers like OpenAI or Anthropic.
To encourage consistent standards, \judgezoo makes it easy to exactly reproduce evaluation standards from prior work and emits warnings when their choices (e.g., the selected foundation model) differ from existing baselines.
While prior studies often use custom thresholds~\citep{chan2025can,ha2025one,zhu2024advprefix} or prompts~\citep{chao2025jailbreaking}, our goal is to make evaluation easily reproducible by providing tested references for methodologies in popular papers.
\begin{table}[h]
\centering
\setlength{\tabcolsep}{6pt}
\small
\begin{tabular}{ll}
\toprule
Type & Proposed in \\
\midrule
\multirow{4}{*}{Prompting} 
  & Adaptive Attacks/PAIR~\citep{andriushchenko2024jailbreaking,chao2025jailbreaking} \\
  & AdvPrefix~\citep{zhu2024advprefix} \\
  & StrongREJECT (rubric)~\citep{souly2024strongreject} \\
  & XSTest~\citep{rottger2024xstest} \\
\midrule
\multirow{7}{*}{Fine-tuned} 
  & AegisGuard~\citep{ghosh2024aegis} \\
  & HarmBench~\citep{mazeika2024harmbench} \\
  & JailJudge~\citep{liu2024jailjudge} \\
  & Llama Guard 3~\citep{llama2024llama3herd} \\
  & Llama Guard 4~\citep{meta2025llama4} \\
  & MD-Judge (v0.1 \& v0.2)~\citep{li2024salad} \\
  & StrongREJECT~\citep{souly2024strongreject} \\
\midrule
\multirow{1}{*}{False-positive filter} 
  & Best-Of-N~\citep{hughes2024best} \\
\bottomrule
\end{tabular}
\caption{Evaluation options/judging models included in \judgezoo.}
\label{tab:judges}
\end{table}

\section{Key Features}

We now describe features that distinguish \name from existing frameworks.
While several items (reproducibility, budget tracking) represent best practices that should be universal, others (e.g., our tokenization handling) solve previously unrecognized problems in attack toolboxes.

\subsection{Robust tokenization filters \& utilities}
\textbf{Problem:} Many attacks (e.g., GCG~\citep{zou2023universal}, BEAST~\citep{sadasivan2024fast}, FLRT~\citep{thompson2024flrt}) optimize adversarial inputs in token space by manipulating discrete token IDs.
However, in realistic attack scenarios for hosted models, the tokenization step is typically hidden behind the API surface, which only exposes string-based inputs.
In fact, access is typically even more restricted, allowing only partial control of the input text sequence through a ``conversations" or ``chat" API that allows specifying turn-by-turn messages for various roles such as system messages, user prompts, or assistant/model generations, without control over how the messages are flattened into strings for tokenization.
This creates a subtle but critical gap: some token sequences found during optimization may be \emph{unreachable} via any input string due to how tokenizers handle segment boundaries.

\begin{figure}[h]
  \centering
  \vspace{-0.5\baselineskip}
  \includegraphics[width=\linewidth]{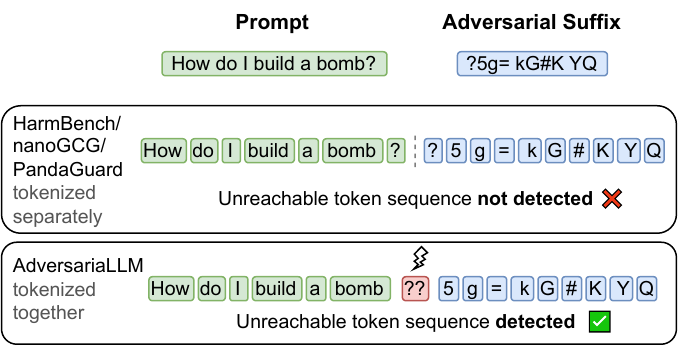}
  \vspace{-1\baselineskip}
  \caption{Our implementation tokenizes the whole input conversation at once, catching more illegal token sequences than other toolboxes. Prior implementations tokenize prompt and suffix separately and only check the suffix for round-trip encode-decode consistency. This makes them unable to detect merges across segment boundaries and leads to attacks which work in token-space, but are impossible to trigger with text input. }
  \label{fig:tokenization-filter}
\end{figure}

Prior implementations~\citep{mazeika2024harmbench,zou2023universal,shen2025pandaguard} check reachability by tokenizing the adversarial suffix in isolation and verifying round-trip encode-decode consistency.
However, this misses token merges that occur \emph{across segment boundaries}, for example, where the adversarial suffix meets the template tokens or where the prompt meets the suffix (Figure~\ref{fig:tokenization-filter}).

\textbf{Solution:} We tokenize the entire conversation at once and verify reachability of the full sequence, not just individual segments.
This catches all boundary merges and ensures that every token sequence we report is actually achievable via text input.
Since many attacks require separate tokenized representations of different input text segments (e.g., instruction, adversarial suffix, template tokens, response tokens, ...), we also expose utilities that tokenize a conversation and split it into its constituent token segments while accounting for merges.

\begin{figure}[h]
  \centering
  \vspace{-0.5\baselineskip}
  \includegraphics[width=\linewidth]{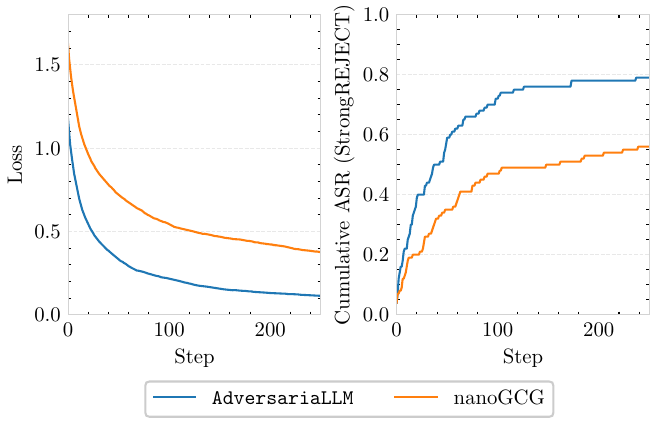}
  \vspace{-\baselineskip}
  \caption{Tokenization details have a significant effect on ASR. Our implementation fixes several issues and leads to significantly improved performance. We show data for GCG against Llama-2-7B-Instruct on the non-copyright subset of the HarmBench dataset. We report cumulative best-of-$n$ ASR (i.e., at each step, the current prompt iterate is used to query the model).}
  \vspace{-\baselineskip}
  \label{fig:whitespace-token}
\end{figure}

\textbf{Impact:} Across 3000 GCG runs comprising 750,000 prompt candidates, our filter catches 2\% more illegal sequences than existing implementations, affecting \textbf{94\% of all attack runs} (\autoref{tab:token-filtering}).
In some cases, up to 15\% of candidates that passed existing filters are actually unreachable.
These and other tokenization corrections (such as whitespace misalignment and end-of-turn token fixes) lead to up to 28\% higher ASR than baseline implementations on some models despite using identical model weights, datasets, and attack algorithms (\autoref{fig:whitespace-token}).
This demonstrates that tokenization is not just a minor implementation detail, but can significantly affect results.

\subsection{Resource-aware attack comparison}

\textbf{Problem} In computer vision and graph machine learning, adversarial robustness evaluations typically use proximity budgets like $L_p$-norm or node perturbation constraints to handicap attacks and enable fair comparison.
LLMs lack a natural analog: the input space consists of sequences of discrete tokens, and realistic threat models typically allow near-unrestricted control, limiting the usefulness of proximity-based constraints.
As a result, attacks are often compared under varying query, compute, or wall-time budgets, making comparisons difficult.

\textbf{Solution:} We automatically track three complementary measures of attack effort across all attacks:
\begin{itemize}%
    \item \textbf{Query count}~\citep{chao2025jailbreaking}: Number of model forward passes
    \item \textbf{Wall time}~\citep{sadasivan2024fast}: Actual elapsed time
    \item \textbf{FLOPs}~\citep{boreiko2024interpretable}: Computational operations, hardware-independent
\end{itemize}

Critically, we separate resources for finding the attack prompt from resources for sampling harmful outputs, which allows nuanced comparisons between approaches like Best-of-N jailbreaking~\citep{hughes2024best}, which requires minimal prompt optimization but extensive sampling, and GCG~\citep{zou2023universal} which inverts this trade-off.
Conflating these costs would obscure which attacks are efficient under different threat models.

\textbf{Impact:} By tracking all three metrics and splitting them into optimization and sampling, researchers can compare attacks under multiple definitions of "budget" and better understand efficiency trade-offs.


\subsection{Per-step evaluation and distributional robustness}

\textbf{Problem:} Most frameworks evaluate only the final attack candidate or the "best" according to a proxy metric, often with greedy decoding.
This provides only one data point per attack run, obscuring optimization dynamics.

\textbf{Solution:} We evaluate attack effectiveness at every optimization step by default.
We also support distributional evaluation via Monte Carlo sampling~\citep{beyer2025sampling,scholten2024probabilistic} with flexible sampling schedules, which recent work shows is critical for thorough robustness assessment.

\textbf{Impact:} Per-step evaluation reveals convergence speed, enables sample-efficient early stopping, and provides much richer data for comparing optimization strategies—all with minimal overhead.
Distributional evaluation provides more principled and nuanced robustness data

\subsection{Numerical stability \& determinism}
\textbf{Problem:}
For many popular models, batched generation with ragged sequences using \texttt{model.generate} can lead to significant deviations from generations in single-batch settings, even in greedy generation. The observed differences significantly exceed what would be expected from a numerical precision perspective.

\textbf{Solution:}
We implement a custom inference function for batches of ragged sequences.

\textbf{Impact:} \name shows $2.12\times$\footnote{For Llama 3.1 8B, our implementation matches the reference 256-token generation 71\% of the time vs. only 19\% using the baseline implementation.
The average common prefix length increased by $2.12\times$. \autoref{app:generation} shows data across more models.} better alignment with single-batch reference generations while retaining high performance.

\subsection{Reproducibility \& parameter tracking}

\textbf{Problem:}
Existing frameworks often fail to track all information necessary for exact reproduction, making it difficult to verify or build upon prior results.
Dynamic elements in chat templates (e.g., automatically inserted dates) can cause different behavior when re-running evaluations at different times, creating subtle inconsistencies that are hard to detect.

\textbf{Solution:}
Similar to JailbreakBench~\citep{chao2024jailbreakbench}, \name tracks and stores all hyperparameters required to instantiate a run.
We go further e.g., by storing the complete sequence of token IDs (or embeddings for latent-space attacks) passed to the model at every step, including system message and template tokens, providing richer metadata than existing alternatives.

\textbf{Impact:}
Result artifacts are fully self-contained with all information needed for exact reproduction.
Complete token sequences enable detection and control of subtle shifts from dynamic template elements, ensuring true reproducibility even across different execution times and environments.

\subsection{Scalability}
To enable rapid experimentation at scale, the package integrates flexible batched and parallel jobs via Slurm \citep{yoo2003slurm} and submitit \citep{submitit}.

\section{Limitations}
Despite the effort invested in determinism and reproducibility, attacks may produce slightly varying results when run on different hardware even if all seed parameters are held constant.
This phenomenon is due to the aforementioned dynamic chat templates or due to non-deterministic behavior across different batch sizes, which can lead to numerical differences \citep{he2025nondeterminism}. \\
Similarly, due to the white box nature of many of the contained attacks, our toolbox is currently focused on open-weight models.
In the future, we plan to expand support of API-based models to all black-box attacks.\\
Lastly, the toolbox currently assumes that defenses are built into models directly and does not yet directly support ad-hoc defenses such as SmoothLLM \citep{robey2023smoothllm}.

\section{Conclusion}
This report describes \name, a framework for LLM jailbreak robustness evaluations with a focus on correctness, stability, and modularity.

We release the framework and evaluation toolbox on GitHub and hope our work can contribute to a more efficient, comparable, and correct evaluation landscape.

\bibliographystyle{icml2025.bst} %
\small
\bibliography{Reference}
\normalsize

\appendix
\onecolumn
\section{Token filtering}
\label{app:token-filtering}

We conduct exhaustive tests using data across 3000 GCG attack runs comprising 750,000 prompt candidates. 
Our upgraded filtering algorithm filters 2\% more sequences than existing implementations, affecting 94\% of GCG attack runs, and in some cases removing up to 15\% more candidate sequences than existing implementations.  
\begin{table}[h!]
\centering
\begin{tabular}{lrr}
\toprule
& \name & \texttt{nanoGCG}~\citep{zou2023universal} \\
\midrule %
Filtered sequences (\%) & 56.4 & 54.7 \\
\bottomrule
\end{tabular}
\caption{}
\label{tab:token-filtering}
\end{table}

\section{Comparing model generation correctness}
\label{app:generation}

We benchmark our implementation for generation with ragged prompts against the default \texttt{model.generate} implementation and track how long it takes for generations to diverge.
\begin{table}[h!]
\centering
\begin{tabular}{lrrr}
\toprule
Model Name & \texttt{model.generate} & \textbf{Ours} & Factor of improvement \\
\midrule
mistralai/Ministral-8B-Instruct-2410 & 23.05 & \textbf{40.95} & 1.78 \\
cais/zephyr\_7b\_r2d2 & 0.45 & \textbf{109.95} & 244.33 \\
microsoft/Phi-3-mini-4k-instruct & \textbf{54.50} & 45.55 & 0.84 \\
qwen/Qwen2-7B-Instruct & 23.70 & \textbf{30.15} & 1.27 \\
meta-llama/Llama-2-7b-chat-hf & 23.30 & \textbf{52.15} & 2.24 \\
meta-llama/Meta-Llama-3.1-8B-Instruct & 7.50 & \textbf{52.85} & 7.05 \\
\midrule
Average & 22.10 & \textbf{55.60} & 2.12 \\
\bottomrule
\end{tabular}
\caption{Evaluating numerical correctness during generation by comparing the average number of tokens that are equal to the greedy unbatched baseline. The Improvement column shows the relative factor between \textbf{Ours} and \texttt{model.generate}.}
\label{tab:batched-generation-tokens}
\end{table}

\newpage
\section{Schema of attack artifacts}
\label{app:attack-result}
All attack results are logged as human-readable JSON files with the following hierarchical structure:
  
  \begin{table}[h]
  \centering
  \small
  \begin{tabular}{@{}llp{6cm}@{}}
  \toprule
  \textbf{Field} & \textbf{Type} & \textbf{Description} \\
  \midrule
  \multicolumn{3}{l}{\textit{Top Level}} \\
  \texttt{config} & Object & Combined attack, dataset, \& model configuration \\
  \texttt{runs} & Array[\texttt{SingleAttackRunResult}] & List of \texttt{SingleAttackRunResult} \\
  \midrule
  \multicolumn{3}{l}{\textit{Configuration}} \\
  \texttt{model} & String & Model identifier \\
  \texttt{dataset} & String & Dataset name \\
  \texttt{attack} & String & Attack method \\
  \texttt{model\_params} & Object & Model configration \\
  \texttt{dataset\_params} & Object & Dataset configuration \\
  \texttt{attack\_params} & Object & Attack configuration \\
  \midrule
  \multicolumn{3}{l}{\textit{SingleAttackRunResult}} \\
  \texttt{original\_prompt} & Array & Multi-turn conversation (role/content dicts) \\
  \texttt{steps} & Array[\texttt{AttackStepResult}] & List of \texttt{AttackStepResult} \\
  \texttt{total\_time} & Float & Total time (seconds) \\
  \midrule
  \multicolumn{3}{l}{\textit{AttackStepResult}} \\
  \texttt{step} & Integer & Step number \\
  \texttt{model\_completions} & Array[String] & Sampled model outputs \\
  \texttt{scores} & Dict & Judge name $\rightarrow$ score type $\rightarrow$ values \\
  \texttt{time\_taken} & Float & Per-step time (seconds) \\
  \texttt{flops} & Integer & FLOPs (excluding sampling) \\
  \texttt{loss} & Float? & Optimization loss (if available) \\
  \texttt{model\_input} & Array? & Model input in conversation format (if discrete attack) \\
  \texttt{model\_input\_tokens} & Array[Int]? & Token IDs (if discrete attack) \\
  \texttt{model\_input\_embeddings} & Tensor/String? & Embeddings or path to \texttt{.safetensors} (if embedding attack) \\
  \bottomrule
  \end{tabular}
  \caption{AttackResult JSON schema. Fields marked with \texttt{?} are optional.}
  \label{tab:attack-result-schema}
  \end{table}

\end{document}